%% file: emnlp2018.tex
\documentclass[11pt,a4paper]{article}
\usepackage[hyperref]{emnlp2018}
\usepackage{times}
\usepackage{tikz}
\usepackage{latexsym}
\usepackage{multirow}
\usepackage{tabularx}
\usepackage{arydshln}
\usepackage{amsmath}
\usepackage{array}
\usepackage{booktabs}
\usepackage{soul}
\usepackage{enumitem}
\usepackage{graphicx}
\usepackage{nicefrac}
\usepackage{url}
\usepackage{subcaption}
\usepackage{natbib}

\newcolumntype{L}[1]{>{\raggedright\let\newline\\\arraybackslash\hspace{0pt}}m{#1}}
\newcolumntype{u}[1]{>{\raggedright\let\newline\\\arraybackslash\hspace{0pt}}p{#1}}
\newcolumntype{C}[1]{>{\centering\let\newline\\\arraybackslash\hspace{0pt}}m{#1}}
\newcolumntype{R}[1]{>{\raggedleft\let\newline\\\arraybackslash\hspace{0pt}}m{#1}}
\newcolumntype{t}[1]{>{\raggedleft\let\newline\\\arraybackslash\hspace{0pt}}p{#1}}

\usetikzlibrary{positioning,arrows,shapes,shapes.multipart,shapes.geometric,fit}
\pgfdeclarelayer{back}
\pgfsetlayers{back,main}

\newcommand{\tikzscale}[0]{0.8}

\aclfinalcopy 

\title{Effective Parallel Corpus Mining using Bilingual Sentence Embeddings}

\author{
Mandy Guo\textsuperscript{$a$}\thanks{\hspace{2mm}equal contribution},
Qinlan Shen\textsuperscript{$b$}\thanks{\hspace{2mm}Work done during an internship at Google AI.} \footnotemark[1],
Yinfei Yang\textsuperscript{$a$}\footnotemark[1], 
Heming Ge\textsuperscript{$a$},
Daniel Cer\textsuperscript{$a$}, 
\\ \rm\textbf{
Gustavo Hernandez Abrego\textsuperscript{$a$}, 
Keith Stevens\textsuperscript{$a$},
Noah Constant\textsuperscript{$a$},
}
\\ \rm\textbf{
Yun{-}Hsuan Sung\textsuperscript{$a$},
Brian Strope\textsuperscript{$a$},
Ray Kurzweil\textsuperscript{$a$}
} \AND
  {\rm\textsuperscript{$a$}Google AI}\\Mountain View, CA, USA \And
  {\rm\textsuperscript{$b$}Carnegie Mellon University}\\Pittsburgh, PA, USA
}

\date{}

\begin{document}
\maketitle
\begin{abstract}
This paper presents an effective approach for parallel corpus mining using bilingual sentence embeddings.
Our embedding models are trained to produce similar representations exclusively for bilingual sentence pairs that are translations of each other. This is achieved using a novel training method that introduces hard negatives consisting of sentences that are not translations but that have some degree of semantic similarity.
The quality of the resulting embeddings are evaluated on parallel corpus reconstruction and by assessing machine translation systems trained on gold vs.\ mined sentence pairs. 
We find that the sentence embeddings can be used to reconstruct the United Nations Parallel Corpus~\cite{uncorpus} at the sentence level with a precision of 48.9\% for en-fr and 54.9\% for en-es. When adapted to document level matching, we achieve a parallel document matching accuracy that is comparable to the significantly more computationally intensive approach of \newcite{jakob2010}. Using reconstructed parallel data, we are able to train NMT models that perform nearly as well as models trained on the original data (within 1-2 BLEU). 

\end{abstract}

\section{Introduction} \label{sec:intro}
Volumes of quality parallel training data are critical to neural machine translation (NMT) systems. 
While large distributed systems have proven useful for mining parallel documents~\cite{jakob2010,antonova2011}, these approaches are computationally intensive and rely on heavily engineered subsystems.
Recent work has approached the problem by training lightweight end-to-end models based on word and sentence level embeddings~\cite{gregoire2017,BOUAMOR18,schwenk2018filtering}. 

We propose a novel method for training bilingual sentence embeddings that proves useful for sentence level mining of parallel data. Sentences are encoded using Deep Averaging Networks (DANs)~\cite{iyyer-EtAl:2015:ACL-IJCNLP}, a simple bag of n-grams architecture that has been shown to provide surprisingly competitive performance on a number tasks including sentence classification~\cite{iyyer-EtAl:2015:ACL-IJCNLP,cer2018}, conversation input-response prediction~\cite{yang2018}, and email response prediction~\cite{henderson2017}. Separate encoders are used for each language with candidate source and target sentences being paired based on the dot-product of their embedded representations. Training maximizes the dot-product score of sentence pairs that are translations of each other at the expense of sampled negatives. We contrast using random negatives with carefully selected hard negatives that challenge the model to distinguish between true translation pairs versus non-translation pairs that exhibit some degree of semantic similarity.  

The efficiency of the sentence encoders and use of a dot-product operation to score candidate sentence pairs is well suited for parallel corpus mining. Efficient encoders reduce the amount of computational resources required to obtain sentence embeddings for a large collection of unpaired sentences. Once the sentence embeddings are available, efficient nearest neighbour search \cite{ann,faiss} can be used to identify candidate translation pairs.

The language pairs English-French (en-fr) and English-Spanish (en-es) are used in our experiments.  Our results show that introducing hard negative sentence pairs that are semantically similar but that are not translations of each other systematically outperforms using randomly selected negatives.  
Our method can be used to reconstruct the United Nations Parallel Corpus~\cite{uncorpus} at the sentence level with a level of precision of 48.9\% P@1 for en-fr and 54.9\% P@1 for en-es. When we adapt our method to document level pairings we achieve a matching accuracy that is comparable to that of the much heavier weight and computationally intensive approach of \newcite{jakob2010}. Training an NMT model using the reconstructed corpus results in models that perform nearly as well as those trained on the original parallel corpus (within 1-2 BLEU). Finally, our method has a modest degree of correlation with the pair quality scores provided by Zipporah \cite{zipporah}. However, our method has higher agreement with human judgments and results in NMT systems with higher BLEU scores when our approach is used to filter the ParaCrawl corpus.

\section{Approach}
This section introduces our bilingual sentence embedding model and the translation candidate ranking task we use for training. Our method for selection of hard negative sentence pairs that are not translations of each other but have some degree of semantic similarity is then presented. Finally, we detail the use of our bilingual sentence embeddings to search for sentences that are translations of each other as well as an adaptation to the matching process to parallel documents. 

\subsection{Translation Candidates Ranking Task}
Given a pair of sentences that are translations of each other $x$ and $y$, the translation candidate ranking task attempts to rank the true translation $y$ over all other sentences, $\mathcal{Y}$, in the given language. This can be accomplished by modeling the translation probability distribution $P(y \mid x)$. Provided with a scoring function $\phi$ that assesses the compatibility between $x$ and $y$, the distribution can be expressed as the following log-linear model:

\begin{equation}
\label{eq:bayes}
P(y \mid x) = \frac{e^{\phi(x, y)}} {\sum_{\bar{y} \in \mathcal{Y}} e^{\phi(x, \bar{y})}}
\end{equation}

To avoid summing over all possible target sentences,
the normalization term is approximated by summing over the compatibility score for matching $x$ to $K-1$ sampled negatives together with the compatibility score for the positive candidate:

\begin{equation}
\label{eq:approx}
P_{approx}(y \mid x) = \frac{e^{\phi(x, y)}} {\sum_{k=1}^{K} e^{\phi(x, y_k)}}
\end{equation}

This formulation is similar to early work on discriminative training of log-linear translation decoding models~\cite{Och:2002}. However, rather than using a weighted sum of manually engineered features, we define $\phi$ to be the dot-product of sentence embeddings for the source, $\mathbf{u}$, and target, $\mathbf{v}$, with \mbox{$\phi(x,y) = \mathbf{u}^\top \cdot \mathbf{v}$}. A similar log-linear sentence embedding based formulation of $P(y|x)$ has been previously used to model conversation and e-mail response prediction~\cite{henderson2017,yang2018}.

\subsection{Bilingual Sentence Embeddings}

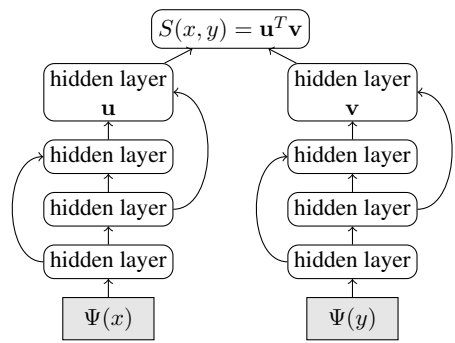
\begin{figure}[htbp]
  \centering
  \input{dual_encoder.tex}
  \caption{\label{fig:dual_encoder}
  \small
    Dual-encoder architecture, where a group of hidden layers encodes source sentence $x$ to $\mathbf{u}$ and a separate group encodes
    target sentence $y$ to $\mathbf{v}$ such that the score $\phi(x,y)$ is the dot-product
    ${\mathbf{u}}^T \cdot \mathbf{v}$.
  }
\end{figure}

Bilingual sentence embeddings are obtain using the dual-encoder architecture illustrated in Figure \ref{fig:dual_encoder}. We use Deep Averaging Networks (DANs)~\cite{iyyer-EtAl:2015:ACL-IJCNLP} to compute sentence-level embedding vectors by first averaging word and bi-gram level embeddings, denoted as $\Psi(x)$ and $\Psi(y)$, for the source and target sentences, respectively.\footnote{Our implementation sums the word and bi-gram embeddings and then divides the result by $sqrt(n)$, where $n$ is the sentence length. The intuition behind dividing by $sqrt(n)$ is as follows: We want our input embeddings to be sensitive to length.
However, we also want to ensure that, for short sequences, the relative differences in the representations are not dominated by sentence length effects.} The word and bi-gram level embeddings are not pretrained but are rather learned during training of the sentence encoders. The averaged representation is provided to a feedforward deep neural network (DNN). Across hidden layers we include residual connections with a skip level of $1$. The final bilingual sentence embeddings are $\mathbf{u}$ and $\mathbf{v}$, which are taken from the last layer of the source and target encoders, respectively. The dot-product of the sentence embeddings, $\mathbf{u}^T \cdot \mathbf{v}$, is used to compute the translation score, $\phi(x,y)$.  

\begin{table*}
\small
\centering
    \begin{tabular}{|c|c|c l|}
    \hline
         & \textbf{Source (Target)} &  & \textbf{Negatives}\\  \hline
          \multirow{8}{*}{en-fr} & & \multirow{2}{*}{\textbf{Random}} & Sa respiration devient laborieuse \\
          & How to display and access shared files & & Benoit Faucon Lieu London \\
          \cdashline{3-4}
          & (Comment afficher et acc\'eder aux fichiers partag\'es) & \multirow{2}{*}{\textbf{Hard}} & Acc\`es à l'environment des fichiers partag\'es\\
          &  & & Des \'el\'ements comme des fichiers de dossiers \\ 
          \cline{2-4}
          & & \multirow{2}{*}{\textbf{Random}} & RCS 871, o\`u le juge Fauteux explique \\
          & The General Delegation for Armaments & & Avis sur les h\^otels \\
          \cdashline{3-4}
          & (La d\'el\'egation g\'en\'erale pour l'armement)& \multirow{2}{*}{\textbf{Hard}} & La 9e arm\'ee , command\'ee par le g\'en\'eral Foch\\
          & & & La d\'el\'egation militaire hongroise compos\'ee de ...\\  \hline  \hline
          \multirow{8}{*}{en-es} & & \multirow{2}{*}{\textbf{Random}} & Alquiler mensual desde : 890 USD \\
          & Oil and gas investments & & \textquestiondown Qu\'e m\'as se deja para preguntar? \\
          \cdashline{3-4}
          & (Inversiones en petr\'oleo y gas) & \multirow{2}{*}{\textbf{Hard}} & Petr\'oleo y gas\\
          &  & & Petr\'oleo y Gas Petroqu\'imica p\'agina \\ 
          \cline{2-4}
          & & \multirow{2}{*}{\textbf{Random}} & Ve el perfil completo de Fleishman \\
          & In Spain, it has clearly chosen the gratuity & & Le\'on de monta\~na en roca \\
          \cdashline{3-4}
          & (En Espa\~na, se ha elegido claramente la gratuidad) & \multirow{2}{*}{\textbf{Hard}} & Dejar propina es una costumbre chilena \\
          & & & Este es un t\'ipico restaurante espa\~nol de Espa\~na \\
          \hline
    \end{tabular}
    \caption{Example of random negatives and hard negatives for en-fr and en-es.}
    \label{tab:negative_examples}
\end{table*}

\begin{figure}[htbp]
  \centering
  \includegraphics[width=.35\textwidth]{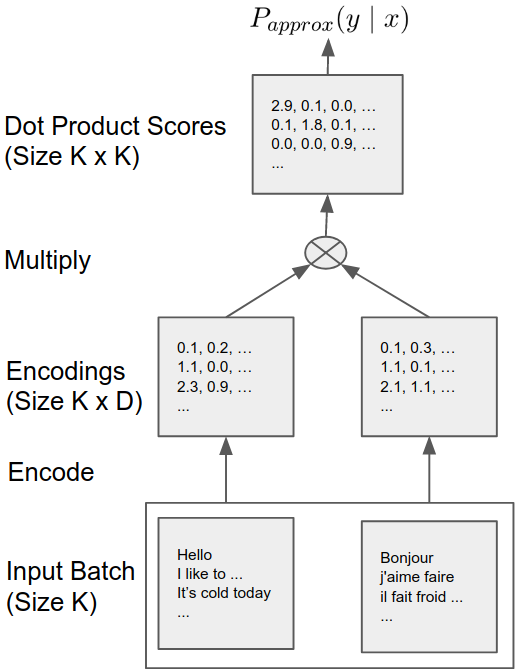}
  \caption{\small
  Matrix multiply trick for dot product model with random sampling.
  }
  \label{fig:matrix_multiply}
\end{figure}

The dual-encoders are trained to the translation candidate ranking task by maximizing the log likelihood of $P_{approx}$. This objective is particularly well suited for mini-batch training. As illustrated in Figure \ref{fig:matrix_multiply}, within a batch, each source and target translation pair servers as a positive example for that particular pairing with alternative pairings within the same batch being treated as negative examples. Given an ordered collection of embeddings for source and target translation pairs, all of the dot-product scores necessary to compute $P_{approx}$ can be determined using a single matrix multiplication of the encoding matrices, $\mathbf{U}$ and $\mathbf{V}^\top$.\footnote{The encoding matrices are composed of the ordered sentence embeddings for all of the source and target sentences within a batch, $\mathbf{U} = (\mathbf{u}_0, \mathbf{u}_1, ..., \mathbf{u}_{k-1}$) and $\mathbf{V} = (\mathbf{v}_0, \mathbf{v}_1, ..., \mathbf{v}_{k-1}$). } After the matrix multiplication the scores assigned to true translation pairs can be found on the diagonal while the scores for incorrect pairings are off-diagonal.

Within our experiments, models differ in their selection of the $K-1$ sampled negatives. Our preliminary models make use of the random sampling strategy that has been proven successful in prior work \cite{henderson2017,yang2018}. Using this strategy consists of randomly composing batches of translation pairs and using the matrix multiplication approach described above to obtain within batch negatives for each incorrect pairing. We employ random shuffling during training resulting in different random negatives for each $\mathbf{u}_i$ across epochs.
As described below we also explore introducing additional hard negatives. This is achieved by extending the target embeddings matrix $\mathbf{V}$ with the sentence embeddings for the hard negatives, which introduces additional off-diagonal values within the matrix of dot-product scores.

\subsection{Semantically Similar Hard Negatives}

As illustrated in Table \ref{tab:negative_examples}, randomly selected negatives results in many pairings that are obviously incorrect without requiring a careful assessment of whether the source and randomly sampled targets are true translations.

Within a mini-batch, the model could likely achieve a reasonable level of performance by simply identifying which source and target sentences are on the same topic or are otherwise semantically related. However, when mining for parallel data, extracting sentence pairs that are not translations of each other but that are rather merely topically related is expected to harm downstream MT systems that train on the erroneous pairs. Given the increased sensitively of NMT models to data quality issues, performance might be harmed even including semantically similar sentences with sufficient differences in meaning between them.\footnote{e.g., add or removing of important details according to the sentence similarity scale proposed by \newcite{agirre-EtAl:2012}.} 

We improve the mining of true translation pairs by making model training more challenging through the introduction of \textit{hard negatives} -- semantically similar translations that are close but not quite identical to the correct translation. The hard negatives are selected using a baseline model trained with randomly sampled negatives. For each source sentence, we identify $M$ hard negatives with target embeddings that achieve high dot-product scores with the source sentence embedding but that are not the correct translation. 
Examples of hard negatives extracted using the baseline model are provided in Table \ref{tab:negative_examples}. Compare to the random negatives, hard negatives are semantically more similar to the correct target translation.

As described above, the hard negatives are appended to the target embedding matrix $\mathbf{V}$. Therefore, instead of training with $K$ candidates, each translation input will be compared with $K+K*M$ candidates, where $K$ is the batch size. In practice, getting hard negatives for the entire dataset is very time consuming.
We only obtain hard negatives for $20\%$ of the data and use random negative sampling for the remainder of the training set.

\subsection{Mining Parallel Data}

One approach to mining parallel data with bilingual sentence level embeddings is to independently pair individual source and target candidates based on the similarity of their embeddings. Prior work that explored using this approach found that the resulting mined sentence pairs produced poor BLEU scores when used for MT training unless they were combined with traditional human translated corpora with known alignments \cite{schwenk2018filtering}. We explore both sentence level and document level mining of parallel corpora. For document level mining, we introduce a novel selection criteria that takes into account the confidence of sentence alignments within a document and sentence position information.

\subsubsection{Document Matching}
\label{sec:doc_match}
Parallel documents are identified as follows: For a given source document, we first run an approximate nearest neighbor (ANN) search for each sentence in the document. This gives us $N$ target sentences for each source sentence (ranked in order of closest match). Let $Y$ be the bag of all target sentences that appear as a match for at least one source sentence. Then for each sentence in $Y$, we look up the document from which they came. We score each candidate document using Eq \ref{eq:doc_match_sent_offset}.\footnote{Selecting the target document that appears the most in $Y$ should give us a rough idea of which target document is most likely to be the translation of a source document. However, this approach is quite naive since we are ignoring many pieces of information: 1. The rank at which each target sentence appeared, 2. The dot product score between the target sentence and the source sentence, and 3. The indices of the target sentence and the source sentence (i.e. the position of the sentences within their respective documents). Since the first two factors indicate the model's confidence in the sentence match, it seems desirable to somehow incorporate this information into our scoring of document matches.} This scoring function takes into account the sentence level nearest neighbor rank of the match for source sentence $x$ to target sentence $y$ in the document being scored, $r(x,y)$. The match rank is linearly combined with a normalized confidence score, $f_1(x,y)$, for the match between $x$ and $y$ as well as the absolute difference between the sentence position index of the source and target sentences, $f_2(x, y)$. The sum of the scoring terms is weighted by the hyperparameters, $w_1$ and $w_2$.

\begin{equation}
\label{eq:doc_match_sent_offset}
 \sum_{y \in D \cap Y}{-r_(x,y) + w_1 * f_1(x, y) + w_2 * f_2(x, y)}
\end{equation}

\subsubsection{Calibrated Confidence Score} \label{sec:score_normalization}

The raw dot product score, $\phi(x,y)$, is a poor choice for the confidence score, $f_1(x,y)$. The score from $\phi(x,y)$ provides a relative metric of a translated sentence's match quality with respect to the source sentence, 
but it is not a globally consistent measurement of how good a translation pair is. For example, scores are not necessarily in the same range nor do they have comparable relative values for different input source sentences. As a result, if we choose $\phi(x,y)$ to score confidence, there is no single threshold we can use to filter out bad results.

\begin{figure}[htbp]
  \centering
  \input{scoring_model.tex}
  \caption{\label{fig:scoring_model}
  \small
    Scoring model based on dual-encoder architecture.
  }
\end{figure}
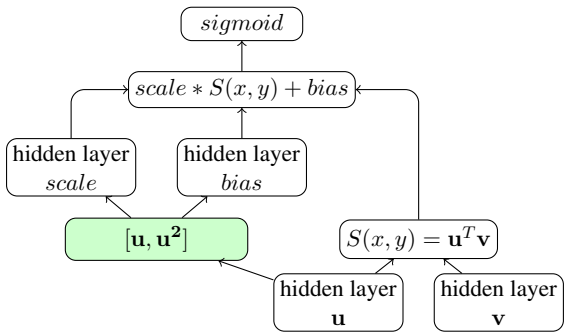

In order to obtain more consistent confidence scores, we propose a novel score normalization model based on dynamic scaling and shifting of the dot product scores. As illustrated in Figure \ref{fig:scoring_model}, the dynamic scaling and shifting values are computed from the source embedding, $\mathbf{u}$, and a pointwise squaring of the values within the source embedding, $\mathbf{u}^2$. The vectors $\mathbf{u}$ and $\mathbf{u}^2$ are concatenated. The scale and bias terms are computed as a weighted sum of the concatenated vectors values. After the dynamic scaling and bias terms are used to calibrate the dot-product score, the resulting calibrated dot-product is passed to a sigmoid in order to obtain a final confidence value between 0 and 1. The weights used to compute the scale and bias terms are trained on held out supervised data.

It is worth noting that because the hidden layers for $scale$ and $bias$ only use features from the source embeddings, it will not affect the ranking of targets.
Thus, we still always use dot-product similarity, $\phi(x,y)$, to retrieve targets via nearest neighbor search. For document level matching, we convert the dot-product values into the calibrated confidence scores, $f_2(x, y)$, without needing to reinspect the target embeddings.

\section{Experiments}

We train our proposed model on two language pairs: English-French (en-fr) and English-Spanish (en-es).
First, we evaluate the performance on the translation candidate ranking task, 
comparing the dual-encoder architectures with random negative sampling versus using hard negatives. 
Then, we present results for document level matching using \newcite{jakob2010}'s method as a strong baseline. We explore training NMT systems using our method to both filter and re-construct parallel corpora. Finally, we assess the level or agreement between our method and human judgments.

\subsection{Data}\label{sec:train_data}
For training the model, we construct a parallel corpus using a system similar to the approach described in \newcite{jakob2010}.
The final constructed corpus contains around 600M en-fr pairs and 470M en-es pairs. To assess the quality of the parallel corpus, we ask human annotators to manually evaluate the constructed pairs. The human annotators judge whether 200 randomly selected sentence pairs for both en-fr and en-es are GOOD or BAD translations. We find that the GOOD translation rate is around $80\%$ for both language pairs. The constructed parallel corpus is split into two parts: a training set ($90\%$) and a held-out dev set ($10\%$), with the held-out dev set being used for our preliminary reconstruction experiments. 

The UN corpus \cite{uncorpus} is used for additional corpus reconstruction experiments. The corpus consists of 800k manually translated UN documents from 1990 to 2014 for the six official UN languages.
86k of these documents are fully aligned at the sentence-level for all 15 language pairs.
We make use of the fully aligned en-fr and en-es document pairs and extract all aligned sentence pairs from those document pairs.
There are a total of 11.3 million aligned sentence pairs each for en-fr and en-es.
Assuming that we have no knowledge about which documents and sentences are aligned, the task is to reconstruct the document and sentence pairs.

We evaluate trained translation models on wmt13~\cite{wmt13} and wmt14~\cite{wmt14} for en-es and en-fr, respectively. Translation models are trained using data taken from the parallel corpus described above that was constructed using \newcite{jakob2010}'s method. Additional translation experiments make use of ParaCrawl\footnote{https://paracrawl.eu}, a dataset containing 4 billion noisy translation pairs for en-fr and 2 billion pairs for en-es. Within Paracrawl, each pair contains pre-computed scores by Zipporah~\cite{zipporah} and the Bicleaner tool, which estimates the translation quality of the pair. We make use of the Zipporah scores to compare translation models trained on filtered versions of the corpus selected using Zipporah versus our method.

\subsection{Experimental Configuration}\label{sec:experimental_config}
Model configuration and hyperparameters for our sentence embedding models are set mostly based on defaults taken from prior work with very minimal tuning on the held-out dev set.
For each language, we build a vocabulary consisting of 200 million unigram and 200 million bi-gram tokens.
All inputs are tokenized and normalized before being fed to the model.
We employ an SGD optimizer with a batch size of 128.
The learning rate is set to 0.01 with a learning decay of 0.96 every 5 million steps.
We train for 50 million steps.

For each encoder layer, we employ a four-layer DNN model which contains 320, 320, 500 and 500 hidden units for each layer respectively.
We apply a ReLU activation in the first three layers and no activation in the final layer. 
We enable residual connections between layers with a skip level of 1.
There is no parameter sharing between the source and target encoder layers.
The size of the unigram and bi-gram embeddings is set to 320 and the embeddings are updated during the training process.
The sentence embedding size is set to 512 for both source and target languages.

The calibrated confidence score is trained jointly with the translation candidate ranking task but with a stop gradient that prevents the confidence task from modifying the bilingual sentence encoders. 
The tasks are trained in a multitask framework with multiple workers, where 90\% of the workers optimize the translation candidate ranking task and the remaining 10\% optimize the confidence task.
We use the same configuration for confidence as when training the translation candidate ranking task. Both use the same batch size 128,
meaning there is 1 positive and 127 negative candidates selected for each pass over an example.
We apply a dropout of 0.4 before feeding the feature vector $[\mathbf{u}, \mathbf{u^2}]$ into the hidden layers that calculate $scale$ and $bias$.

\subsection{Dev Set Sentence Level Matching}

\begin{table*}
\centering
    \begin{tabular}{c | c c c || c c c } 
        \hline
        \multirow{2}{*}{Negative Selection Approach} & \multicolumn{3}{c||}{en-fr} & \multicolumn{3}{c}{en-es} \\
        \cline{2-7}
        &  P@1 & P@3 & P@10  &  P@1 & P@3 & P@10 \\ 
        \hline
        Random Negatives                  & 70.49 & 80.03 & 86.39 & 67.81 & 77.37 & 84.42 \\
        Random Negatives (Augmented)      & 70.67 & 79.99 & 86.14 & 70.47 & 79.79 & 86.33 \\
        (5) Hard Negatives                & 78.31 & 85.30 & 89.52 & 73.46 & 82.37 & 87.75 \\
        (10) Hard Negatives               & 77.06 & 84.04 & 88.70 & 74.92 & 83.29 & 88.14 \\
        (20) Hard Negatives               & 78.29 & 85.06 & 89.58 & 74.84 & 82.86 & 88.23 \\
        \hline
    \end{tabular}
\caption{Precision at N (P@N) results on the evaluation set for models built using the random negatives and ($M$) hard negatives. Models attempt to select the true translation target for a source sentence against 10M randomly selected targets.}
\label{tab:target_10M}    
\end{table*}

We first evaluate the trained models on the translation target retrieval task and use precision at N (P@N) as our evaluation metric.
For every source sentence in the dev set, we run the model and find the nearest neighbors from a set of possible target sentences. 
Previous work~\cite{henderson2017,yang2018} usually evaluated P@N from 100 examples (1 positive and 99 negatives).
We find that this does not work well for the translation target ranking task.
Rather, the P@N of 100 metric goes up to 99.9\% quickly and provides no differentiation between models trained with different configurations.

In this work, we evaluate the P@N from the true target sentence (positive) and 10 million random selected target (negatives) given a source sentence.
We score all selected targets using the translation pair scoring model and rank them accordingly.
The P@N score evaluates if the true translation target (positive) is in the top N target candidates.
We evaluate the model with random sampling and $M$ hard negatives for $M$=5, 10, 20.
Recall that the number of negatives is equal to the batch size for the models trained with random sampling.
The number of negatives for hard negative models, however, is $K+ K*M$ where $K$ is the batch size.
To make a fair comparison, we also evaluate a model trained with additional random samples, by augmenting the number of random negatives to $K+ K*20$.

Table \ref{tab:target_10M} shows the P@N results of the proposed models for N=1, 3, 10.
The model with random negatives provides a strong baseline for finding the right translation target, 
with a P@1 metric of 70.49\% for en-fr and 67.81\% for en-es.
The augmented random negative model performs better than the base random negative model for en-es.
However, the hard negative models outperform the random negative models across all metrics.
Even with only 5 hard negatives, the P@1 metrics improved by 8\% for en-fr and 3\% for en-es.
The addition of more hard negatives, however, does not always further improve performance.

\section{Reconstructing the United Nations Corpus}
In this section, we demonstrate that the proposed model can be used to efficiently reconstruct the United Nations (UN) Parallel Corpus~\cite{uncorpus}.

\subsection{UN Sentence Level Matching}
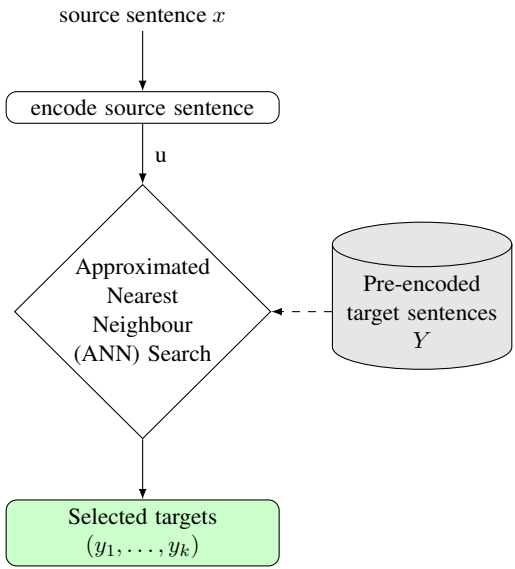
\begin{figure}[htbp]
  \centering
  \input{un_pipeline.tex}
  \caption{\label{fig:un_pipeline}
  \small Target sentence retrieval pipeline.
  }
\end{figure}

We first apply the dual-encoder model to mine target candidates at the sentence level.
As mentioned in section \ref{sec:intro}, one of the advantages of the dual-encoder model is that it is straightforward to use it to encode the source and target sentences separately.
Taking advantage of this property, we first pre-encode all target sentences into a target database, and then we iterate through the source sentences to retrieve the potential targets for each one of them using an approximated nearest neighbour (ANN) search \cite{ann}.
The target sentence retrieval pipeline using ANN search is shown in Figure \ref{fig:un_pipeline}.

Once again we first use P@N as the evaluation metric for target retrieval, for N=1, 3, 10.
We evaluate the two random sampling models and a hard negative model with 20 hard negatives for each example.
As shown in table \ref{tab:target_UN}, with random negatives, the P@1 metric is 34.83\% for en-fr and 44.89\% for en-es.
Adding hard negatives boosts the performance on all metrics, improving the P@1 metric more than 10\% absolute in both en-fr and en-es -- 48.9\% for en-fr and 54.9\% for en-es.

\begin{table*}
\centering
    \begin{tabular}{c | c c c || c c c } 
        \hline
        \multirow{2}{*}{Negative Selection Approach} & \multicolumn{3}{c||}{en-fr} & \multicolumn{3}{c}{en-es} \\
        \cline{2-7}
        &  P@1 & P@3 & P@10  &  P@1 & P@3 & P@10 \\ 
        \hline
        Random Negative                  & 34.83 & 47.99 & 61.20 & 44.89 & 58.13 & 70.36 \\
        Random Negative (Augmented)      & 36.51 & 49.07 & 61.37 & 47.08 & 59.55 & 71.34 \\
        (20) Hard Negative               & 48.90 & 62.26 & 73.03 & 54.94 & 67.78 & 78.06 \\
        \hline
    \end{tabular}
\caption{Precision at N (P@N) of target sentence retrieval on the UN corpus. Models attempt to select the true translation target for a source sentence from the entire corpus (11.3 million aligned sentence pairs.) }
\label{tab:target_UN}    
\end{table*}

\subsection{UN Document Level Matching}

In our final reconstruction experiment, we make use of the document level matching method outlined in section \ref{sec:doc_match}.  $N$ is set to 10 based on prior experiments with the translation matching  task on the dev set. The hyperparameters $w_1$ and $w_2$ are set to $5$ and $-2$, respectively. 
We compare using the document matching score proposed by Eq. (\ref{eq:doc_match_sent_offset}) to scoring document pairs by counting the number of Viterbi aligned sentences linking the two together. As a strong baseline, we also include the application of \newcite{jakob2010}'s method to the UN dataset. 

Table \ref{tab:document_matching_accuracy} shows the document matching accuracies. Using Eq. (\ref{eq:doc_match_sent_offset}) to score document matches outperforms counting mutually aligned sentences. Moreover, while our approach is simpler and less computationally intensive than \newcite{jakob2010}'s, it obtains a promising level of performance.

\begin{table}
\centering
    \begin{tabular}{|c|c|c|}
        \hline
        Matching method & en-fr & en-es \\ \hline
        Alignment Counts & 82.1 & 85.1\\
        Our approach Eq. (\ref{eq:doc_match_sent_offset}) & 89.0 & 90.4 \\
        \hline
        \newcite{jakob2010} & 93.4 & 94.4\\
        \hline
    \end{tabular}
    \caption{Accuracy of document matching on UN corpus.}
    \label{tab:document_matching_accuracy}
\end{table}

\subsection{Evaluation Using a Translation Model}

As a proof of concept on using our mined translation pairs as training data, we train translation models using original versus mined parallel sentence pairs. The training data is sourced from the UN corpus and we evaluate on wmt13~\cite{wmt13} and wmt14~\cite{wmt14} for en-es and en-fr, respectively, with performance assessed using BLEU \cite{bleu}.

We examine two versions of the reconstructed corpora.
In the first version, we take the highest scoring match at the sentence level as the mined parallel sentence pairs and these pairs are then filtered by their calibrated confidence score\footnote{The confidence model is trained with a dev set which consist of 1/10 of UN corpora, these data are removed from training.} with default threshold 0.5.
In the second version, we perform document level matching over the UN dataset. 
Within paired documents, we follow \newcite{jakob2010} and employ a dynamic programming sentence alignment algorithm informed by sentence length and multilingual probabilistic dictionaries. 
In both versions, we drop sentence pairs where both sides are either identical or a language detector declares them to be in the wrong language. 
As a post-processing step, the resulting translations are resegmented using the Moses tokenizer and true-cased before evaluation \cite{koehn2007}.

We train Transformer based translation models~\cite{vaswani2017attention}  on the reconstructed UN corpora for en-fr and en-es and compare them with models trained on the original UN pairs, which we use as Oracle models. The Transformer translation models make use of  a model dimension of 512 and a hidden dimension of 2048, with 6 layers and 8 attention heads. The models use the Adam optimizer with the training schedule described in \citet{vaswani2017attention}. For each language pair, sentence pairs are segmented using a shared 32,000 wordpiece vocabulary \cite{schuster2012}. Sentence pairs are then batched together by approximate sequence length with 128 sentences per batch.
We train each model until convergence (approximately 120K steps).

\begin{table}
    \centering
    \begin{tabular}{|c|c|c|}
        \hline
                        &  \textbf{en-fr} & \textbf{en-es} \\
                        & \textbf{(wmt14)} & \textbf{(wmt13)} \\ \hline
        Mined sentence level    & 29.63 & 29.03 \\
        Mined document level  & 30.05 &  27.09 \\ \hline
        Oracle    & 30.96  &  28.81 \\
        \hline
    \end{tabular}
    \caption{BLEU scores on WMT datasets of the NMT models trained on original UN pairs (Oracle) and on two versions of mined UN corpora.}
    \label{tab:un_wmt}
\end{table}

Table \ref{tab:un_wmt} shows the results obtained from the models trained on the different variations of the parallel data.
The models trained with mined pairs perform very close to the Oracle model,
demonstrating the effectiveness of the proposed parallel corpus mining approach.
Training on the mined sentence level pairs even does slightly better than using the Oracle data for en-es. This is presumably because the mined pairs are cleaner due to the filtering step.
We notice, however, that training on the UN corpus gives translation results that are much lower than the state-of-the-art on the WMT evaluation sets.
This is likely due to the fact that the UN parallel corpus is small and drawn from a particularly restricted domain.

\begin{table}
    \centering
    \begin{tabular}{|c|c|c|}
        \hline
                        &  \textbf{en-fr} & \textbf{en-es} \\
                        & \textbf{(wmt14)} & \textbf{(wmt13)} \\ \hline
        Our data    & 39.81 &  33.75 \\
        Zipporah    & 39.29 &  33.58 \\
        \hline
    \end{tabular}
    \caption{BLEU scores on WMT datasets of the NMT models trained on filtered ParaCrawl data.}
    \label{tab:paracrawl_wmt}
\end{table}

\subsection{Filtered ParaCrawl data}
\label{filteredparacrawl}
We compare the performance of training  translation models\footnote{Using the same model parameters as earlier experiments.} on ParaCrawl data filtered using Zipporah scores versus our scoring method. For this experiment, our confidence score is fine-tuned on the ParaCrawl corpus using an additional 900k positive and 900k negative examples selected based on having extreme Zipporah scores.\footnote{Extreme positive score values from Zipporah  are considered to be those in the top 1\% of the agreement scores found in the ParaCrawl corpus. Extreme negative score values are considered to be agreement scores in the bottom 50\% of the Zipporah scores for ParaCrawl.}
With Zipporah, we select all examples from ParaCrawl with a Zipporah score greater  than  or  equal  to  0, which is the threshold used in the official release. There are 43 million such pairs in en-fr and 24 million in en-es.   We then select the same number of pairs from the ParaCrawl data that have the highest scores from our fine-tuned model. 
As illustrated in Table \ref{tab:paracrawl_wmt}, models trained on our filtered data slightly outperform those trained on data filtered by Zipporah.

The performance achieved by the ParaCrawl trained models on the WMT test data is quite high. This suggests that filtered ParaCrawl data is a good source of general-purpose training material.
We expect further improvement in translation performance when the ParaCrawl extracted pairs are combined with additional WMT training sets.

\section{Analysis}

On the ParaCrawl corpus we find that the Pearson's $r$ between Zipporah and our calibrated confidence scores is only $0.4$. This correlation is quiet low given the level of translation performance achieved by both methods when they are used to select training pairs for an NMT system and suggests that the two methods may provide complementary information.

We assess the agreement of the two methods on extreme score values.\footnote{For this analysis we use the same definition of extreme Zipporah scores as in section \ref{filteredparacrawl}}  We sample a balanced data set consisting of 100k pairs with extreme positive Zipphora values and 100k pairs with extreme negative values. At a threshold of 0.5 and without an fine-tuning, our method agrees with the extreme Zipporah scores with an accuracy of 78.2\% for en-fr and 80.5\% for en-es. However, using the confidence scores fine-tuned to ParaCrawl from section \ref{filteredparacrawl}, we achieve a high level of agreement of 98.4\% for en-fr and 98.6\% with fine-tuning. 

We perform an evaluation using human judgments comparing our scoring model against Zipporah scores on the ParaCrawl data.
As in the filtering experiments, we select all examples from ParaCrawl with a Zipporah score greater than or equal to zero and then select a matching number of pairs with the highest scores from our model. We then sample 200 examples from each set and send them to translation professionals for evaluation. Each example is examined by one annotator that labels the pair as either a GOOD or BAD translation.
A GOOD translation means more than 70\% of a sentence is correctly translated in the paired sentences, meaning most of the information is conveyed.

Table \ref{tab:paracrawl_human_eval} shows the GOOD translation rate for each sampled subset.
The performance between the two approaches is close for en-es and the proposed score normalization model is 4\% better for en-fr.
In our analysis of the BAD translation pairs, one common failure pattern from the proposed model is that one of the sentences is only partially translated in the other sentence.
This is likely because we are still missing enough of these types of hard negatives in the training data.
We also find our model produces more pairs where the sentences on both sides are identical.
These identical pairs are mostly labeled as BAD translations because they are unlikely to be actual translations.

\begin{table}[]
    \centering
    \begin{tabular}{|c|c|c|}
        \hline
        & en-fr & en-es \\ \hline
        zipporah  & 72.0 & 74.0 \\
        our model & 76.0 & 74.5 \\
        \hline
    \end{tabular}
    \caption{GOOD translation rate (\%) annotated by translation professionals.}
    \label{tab:paracrawl_human_eval}
\end{table}

\section{Related Work}
The problem of obtaining high-quality parallel corpora, or bitexts, is one of the most critical issues in machine translation.
One longstanding approach for extracting parallel corpora is to mine documents from the web \cite{resnik1999mining}.
Much of the previous work on parallel document mining has relied on using metadata, such as document titles \cite{yang2002mining}, publication dates \cite{munteanu2005improving,munteanu2006extracting} or document structure \cite{chen2000parallel,resnik2003web,shi2006dom}, to identify bitexts. 

Another direction, however, is to identify bitexts using only textual information, as the metadata associated with documents can often be sparse or unreliable \cite{jakob2010}.
Some text-based approaches for identifying bitexts rely on methods such as n-gram scoring \cite{jakob2010}, named entity matching \cite{do2009mining}, and cross-language information retrieval \cite{utiyama2003reliable,munteanu2005improving}.

There is active research on using embedding-based approaches where texts are mapped to an embedding space in order to determine whether they are bitexts. 
\citet{gregoire2017} use a Siamese network \cite{yin2015abcnn} to map source and target language sentences into the same space, then classify whether the sentences are parallel based on labelled data.
\citet{hassan2018achieving} obtain English and Chinese sentence embeddings in a shared space by averaging encoder states from a bilingual shared encoder NMT system.
The cosine similarity between these sentence embeddings is then used as a measure of cross-lingual similarity between the sentences, which can then be used to filter out noisy sentence pairs.
\citet{schwenk2018filtering} use a similar approach but learn a joint embedding over nine languages.
Our proposed model is differs from previous approaches, as it uses a dual-encoder architecture instead of encoder-decoder architecture.
Not only is the dual-encoder architecture is more efficient~\cite{henderson2017}, it also allows us to directly train toward extracting parallel sentences from a collection of candidates.

\section{Conclusion}
In this paper, we present an effective parallel corpus mining approach using sentence embeddings produced by a bilingual dual-encoder model. 
The proposed model encodes source sentences and target sentences into sentence embeddings separately and then calculates the dot-product score for these two embedding vectors to assess translation pair quality. 
We propose the selection of hard negatives that consist of semantically similar sentence pairs that are not translations of each other. 
Our experiments reveal that using hard negatives improves the ability of our model to identify true translation pairs. We find the proposed method to be useful for both mining and filtering parallel data. Our method compares favorably to Zipporah for filtering, while for mining it provides a lightweight alternative to \newcite{jakob2010}'s method.

\section*{Acknowledgement} 
We thank our teammates from Descartes, Translate and other Google groups for their feedback and suggestions.
Special thanks goes to Wei Wang and Melvin Johnson for extracting translation data and discussions on NMT models.

\bibliography{emnlp2018}
\bibliographystyle{acl_natbib_nourl}

\end{document}

%% file: dual_encoder.tex
\begin{tikzpicture}[scale=\tikzscale, every node/.style={transform shape}, every text node part/.style={align=center}]
  \tikzstyle{layer}=[rectangle, draw=black, minimum height=7mm, inner sep=3pt, fill=white]
  \tikzstyle{input}=[fill=black!10, minimum width=15mm, text height=1.5ex, text depth=.25ex]
  \tikzstyle{hidden}=[rounded corners, minimum width=2cm, minimum height=3mm]
  \tikzstyle{connect}=[->, thin]
  \tikzstyle{bend-left-connect}=[bend left=90, thin]
  \tikzstyle{bend-right-connect}=[bend right=90, thin]

  \node[layer, input] (feature) {$\Psi(x)$};
  \node[layer, input, right=25mm of feature] (response-feature) {$\Psi(y)$};

  \node[layer, hidden, above=3mm of feature](hidden-1-feature) {hidden layer};
  \node[layer, hidden, above=3mm of hidden-1-feature](hidden-2-feature) {hidden layer};
  \node[layer, hidden, above=3mm of hidden-2-feature](hidden-3-feature) {hidden layer};
  \node[layer, hidden, above=3mm of hidden-3-feature](hidden-4-feature) {
    hidden layer \\ $\mathbf{u}$
  };

  \node[layer, hidden, above=3mm of response-feature](hidden-1-resp-feature) {hidden layer};
  \node[layer, hidden, above=3mm of hidden-1-resp-feature](hidden-2-resp-feature) {hidden layer};
  \node[layer, hidden, above=3mm of hidden-2-resp-feature](hidden-3-resp-feature) {hidden layer};
  \node[layer, hidden, above=3mm of hidden-3-resp-feature](hidden-4-resp-feature) {
     hidden layer \\  $\mathbf{v}$
  };
  \node[fit=(hidden-4-resp-feature)(hidden-4-feature)](top-feature) {};
  \node[layer, hidden, above=2mm of top-feature] (logit-feature) {$S(x,y) = {\mathbf{u}}^T  \mathbf{v}$};
  \path
    [connect] (feature) edge (hidden-1-feature)
    [connect] (hidden-1-feature) edge (hidden-2-feature)
    [connect] (hidden-2-feature) edge (hidden-3-feature)
    [connect] (hidden-3-feature) edge (hidden-4-feature)
    [connect] (response-feature) edge (hidden-1-resp-feature)
    [connect] (hidden-1-resp-feature) edge (hidden-2-resp-feature)
    [connect] (hidden-2-resp-feature) edge (hidden-3-resp-feature)
    [connect] (hidden-3-resp-feature) edge (hidden-4-resp-feature)
    [connect] (hidden-4-resp-feature) edge (logit-feature)
    [connect] (hidden-4-feature) edge (logit-feature)
    [bend-left-connect] (hidden-1-feature) edge (hidden-3-feature)
    [bend-left-connect] (hidden-1-resp-feature) edge (hidden-3-resp-feature)
    [bend-right-connect] (hidden-2-feature) edge (hidden-4-feature)
    [bend-right-connect] (hidden-2-resp-feature) edge (hidden-4-resp-feature)
  ;

\end{tikzpicture}

%% file: scoring_model.tex
\begin{tikzpicture}[scale=\tikzscale, every node/.style={transform shape}, every text node part/.style={align=center}]
  \tikzstyle{layer}=[rectangle, draw=black, minimum height=7mm, inner sep=3pt, fill=white]
  \tikzstyle{input}=[fill=black!10, minimum width=15mm, text height=1.5ex, text depth=.25ex]
  \tikzstyle{encode_feature}=[rounded corners, fill=green!20, minimum width=30mm, text height=1.5ex, text depth=.25ex]
  \tikzstyle{hidden}=[rounded corners, minimum width=2cm, minimum height=3mm]
  \tikzstyle{connect}=[->, thin]
  \tikzstyle{connect-round}=[->, thin, rounded corners=5pt]
  \tikzstyle{bend-left-connect}=[bend left=90, thin]
  \tikzstyle{bend-right-connect}=[bend right=90, thin]

  \node[layer, hidden, ](hidden-4-feature) {
    hidden layer \\ $\mathbf{u}$
  };

  \node[layer, hidden, right=5mm of hidden-4-feature](hidden-4-resp-feature) {
     hidden layer \\  $\mathbf{v}$
  };

  \node[fit=(hidden-4-resp-feature)(hidden-4-feature)](top-feature) {};
  \node[layer, hidden, above=2mm of top-feature] (logit-feature) {$S(x,y) = {\mathbf{u}}^T  \mathbf{v}$};
  \node[layer, encode_feature, left=15mm of logit-feature] (input-encode-feature) {$[\mathbf{u}, \mathbf{u^2}]$};
  
  \node[above=7mm of input-encode-feature](placeholder-scale-bias) {};
  \node[layer, hidden, left=2mm of placeholder-scale-bias](input-scale) {
    hidden layer \\ $scale$
  };
  
  \node[layer, hidden, right=2mm of placeholder-scale-bias](input-bias) {
    hidden layer \\ $bias$
  };
  
  \node[layer, hidden, above=5mm of input-bias](score-scaling) {$scale*S(x,y) + bias$};
  
  \node[layer, hidden, above=5mm of score-scaling](score-sigmoid){$sigmoid$};
  
  \draw[connect, rounded corners=5pt] (input-scale) |- (score-scaling);
  \draw[connect, rounded corners=5pt] (logit-feature) |- (score-scaling);
  \path
    
    [connect] (hidden-4-feature) edge (input-encode-feature)
    [connect] (hidden-4-resp-feature) edge (logit-feature)
    [connect] (hidden-4-feature) edge (logit-feature)
    [connect] (input-encode-feature) edge (input-scale)
    [connect] (input-encode-feature) edge (input-bias)
    [connect] (input-bias) edge (score-scaling)
    [connect] (score-scaling) edge (score-sigmoid)
    
  ;

\end{tikzpicture}

%% file: un_pipeline.tex
\begin{tikzpicture}[scale=0.8, every node/.style={transform shape}]

  \tikzstyle{arrow}=[->, thin, -latex]
  \tikzstyle{decision}=[diamond, draw=black, text width=6em, text centered, inner sep=1pt]
  \tikzstyle{box}=[rectangle, draw=black, text width=11em, text centered, rounded corners]
  \tikzstyle{dbase}=[cylinder, draw=black, text width=7em, text centered, shape border rotate=90, aspect=0.25, fill=black!10]

  \node[] (input) {source sentence $x$};
  \node[box, below=of input] (encoder) {encode source sentence};
  \node[decision, below=of encoder] (annsearch) {Approximated Nearest Neighbour (ANN) Search};
  \node[dbase, right=of annsearch] (targetset) {Pre-encoded \mbox{target sentences} $Y$ };
  \node[box, fill=green!20, below=of annsearch] (finaltargets) {\mbox{Selected targets} $(y_1, \ldots , y_k)$};

  \path
    [arrow] (input) edge (encoder)
    [arrow] (annsearch) edge (finaltargets)
  ;
  \path [arrow] (encoder) edge node [near start, xshift=3mm, yshift=-3mm] {u} (annsearch);
  \path
    [arrow, dashed] (targetset) edge  (annsearch)
  ;
\end{tikzpicture}